\documentclass[11pt]{article}
\usepackage{geometry}
\usepackage{coling2020}
\usepackage{times}
\usepackage{url}
\usepackage{latexsym}
\usepackage{microtype}
\usepackage{tabularx}
\usepackage{graphics}
\usepackage{graphicx}
\usepackage{courier}
\usepackage{color}
\usepackage{multirow}
\usepackage[warn]{textcomp}
\usepackage{booktabs}
\usepackage{comment}
\usepackage{wrapfig}
\usepackage{array,makecell}
\usepackage[shortlabels]{enumitem}
\colingfinalcopy 

\title{{\sc MultiSem} at SemEval-2020 Task 3: \\  Fine-tuning BERT for Lexical Meaning}

\author{Aina Gar\'i Soler \\
  Universit\'e Paris-Saclay\\
 CNRS, LIMSI \\
 91400 Orsay, France \\
  {\tt aina.gari@limsi.fr} \\\And
  Marianna Apidianaki \\
  Department of Digital Humanities \\
  University of Helsinki \\
  Helsinki, Finland \\ 
  {\tt marianna.apidianaki@helsinki.fi} \\}

\date{}

\begin{document}
\maketitle
\begin{abstract}
  We present the {\sc MultiSem} systems submitted to SemEval 2020 Task 3: Graded Word Similarity in Context (GWSC).  
  We experiment with injecting semantic knowledge into pre-trained BERT models through fine-tuning  
  on lexical semantic tasks related to GWSC.
  We use existing semantically annotated datasets 
  and 
  propose to approximate similarity through automatically generated lexical substitutes in context.
  We participate in both GWSC subtasks and address two languages, English and Finnish.
  Our best English models 
  occupy the third and fourth positions in the ranking for the two subtasks. Performance is lower for the Finnish models 
  which are mid-ranked in the respective subtasks, 
  highlighting the important role of data availability for fine-tuning.  
\end{abstract}



\blfootnote{
    %
    %
    %
    %
    \hspace{-0.65cm}  
    This work is licensed under a Creative Commons 
    Attribution 4.0 International Licence.
    Licence details:
    \url{http://creativecommons.org/licenses/by/4.0/}
    %
    %
}

\section{Introduction}
\label{sec:intro}

The meaning of words is strongly tied to 
the context in which they occur: Different contexts might point to different senses or indicate subtler meaning nuances. 
SemEval 2020 Task 3  ``Graded Word Similarity in Context'' (GWSC) \cite{task3description}
explores 
the 
effect of context on 
meaning, and 
 proposes to predict the similarity of 
word instances in a continuous, or graded,  
fashion. GWSC is based on the CoSimLex dataset \cite{armendariz-etal-2020-cosimlex} and consists of two subtasks where models have to predict (1) the shift in meaning similarity for a pair of words ($w_{a}$, $w_{b}$) occurring in  different contexts, 
and (2) the similarity of two word instances 
in the same context.  
This is illustrated by 
sentences $c_1$ and $c_2$, 
two contexts where {\it dinner} and {\it breakfast}  co-occur. 

\begin{center}
\parbox{15cm}{
   \begin{itemize} 
       \item[$c_1$] (...) After Mickey rings the \underline{\it dinner} bell, Goofy foolishly leaves the driver's seat for \underline{\it breakfast}. 
       \vspace{-1mm}
    \item[$c_2$] Residence Inns typically feature a complimentary small hot \underline{\it breakfast} in the morning and a free light \underline{\it dinner} or snack reception on weekday evenings (...)   
       \end{itemize} 
}   
\end{center}

\noindent A change in meaning  similarity occurs between the highlighted words in the two sentences
: They 
 are less similar in context $c_1$ where {\it dinner} is part of a noun compound ({\it dinner bell}),  than in context $c_2$ where they describe different kinds of meals.   
The shift in meaning is  reflected in the gold similarity scores assigned to these instance pairs in the GWSC dataset (4.39 vs. 5.35).

We build models for these two subtasks by fine-tuning BERT on existing lexical 
similarity datasets.  Additionally, we propose to approximate similarity of words in context through automatically generated lexical substitutes.  
We build and evaluate models in two languages, English and Finnish. 
In Subtask 1, our English and Finnish models ranked third and sixth out of nine participants. In Subtask 2, they are both found at the fourth position among ten participants.\footnote{Our code will be made available at \url{https://github.com/ainagari/semeval2020-task3-multisem}}


\section{Background}

Our methodology 
draws inspiration from recent work 
on injecting semantic information into pre-trained language models (LMs). This can be done at two stages: during model pre-training or during fine-tuning. \newcite{lauscher2019informing} opt for the first,  
adding an additional lexical task to BERT's two training objectives (language modelling and next sentence prediction) \cite{devlin2019bert}. The semantic knowledge used in this additional task 
comes from pre-defined lexicographic resources (like WordNet \cite{Miller1995}), and is shown to  be  
beneficial on almost all tasks in the  GLUE benchmark \cite{wang-etal-2018-glue}. 

 \newcite{arase-tsujii-2019-transfer} inject semantic knowledge into BERT by  
fine-tuning the pre-trained model on paraphrase data.  
They subsequently fine-tune the model again for the related tasks of  paraphrase identification and semantic equivalence assessment, 
and report results that demonstrate improved performance over a model that has not been exposed to paraphrase data.  
We follow their approach and fine-tune BERT models for English and Finnish on 
a set of semantic tasks that are closely related to the GWSC task, since no training data is available for GWSC. 

One of our tasks is inspired by the 
retrofitting approach of   \newcite{shi-etal-2019-retrofitting}.
This consists in gathering sentence pairs from the Microsoft Research Paraphrase Corpus (MRPC)  \cite{dolan-etal-2004-unsupervised}
that share a word and which are paraphrases of each other (T) or not (F). Shi et al. propose an orthogonal transformation for ELMo \cite{peters2018deep} that is trained to bring representations of word instances closer when they appear in meaning-equivalent contexts. They show that this retrofitting approach improves ELMo's performance in a wide range of semantic tasks at the sentence level (sentiment analysis, inference and sentence relatedness). We follow their data collection method to obtain word instances for fine-tuning BERT. We replace the MRPC with the Opusparcus resource \cite{creutz-2018-open} since it covers two of the languages addressed in GWSC, English and Finnish.


\section{System Overview}

\subsection{Datasets} \label{sec:data}

We fine-tune pre-trained BERT  models on 
semantic tasks that are related to  GWSC. 
We specifically select tasks that address the similarity of word meaning in context, and use 
the corresponding datasets to make BERT more sensitive to this specific aspect of 
meaning. Table \ref{tab:dataset_examples} contains annotated  
instances from each dataset used in our experiments. 

\paragraph{Usim} 
The Usim dataset  
contains 10 sentences 
for each of 56 words of different parts of speech, 
manually annotated with pairwise usage similarity scores  \cite{erketal2009,erk2013measuring}.\footnote{\url{http://www.dianamccarthy.co.uk/downloads/WordMeaningAnno2012/}} 
As in GWSC, similarity scores are graded  
and range from 1 (completely different) to 5 (same meaning). The Usim sentences  come from the SemEval 2007 Lexical Substitution task dataset  \cite{mccarthy2007semeval}.\footnote{\url{http://www.dianamccarthy.co.uk/task10index.html}}
To binarize the usage similarity scores and use them for fine-tuning, we consider 
sentence pairs annotated with low similarity scores (score $<2$) as instances denoting 
a different meaning (F), and highly similar sentence pairs (score $>4$) as instances 
of the same sense (T). In total, we use 1,399  Usim sentence pairs for fine-tuning.

\paragraph{Concepts in Context (CoInCo)} 

The CoInCo corpus  \cite{kremer-etal-2014-substitutes} contains manually selected substitutes for all content words in a sentence. Substitute overlap between different word instances reflects their semantic similarity: instance pairs with similar meaning share a higher number of substitutes.
We binarize the data as in \newcite{gari-soler-etal-2019-word} by assigning instance pairs to a class describing the same (T) or different (F) meaning depending on their shared substitutes.
 The data sample used by Gar\'i Soler et al. 
 contains 
 instances with at least four substitutes: 
 T pairs involve instances that have 
 at least 75\% of 
 substitutes in common, and F pairs correspond to 
 instances with no substitute overlap. 
 We gather additional data from CoInCo 
 by relaxing the class inclusion constraints. Specifically, we 
 retain all instances regardless of the number of available substitutes. We consider as T examples instance pairs 
 that have at least 50\% of 
 substitutes in common, and as F examples  
 pairs that share at most one substitute.

We retain up to 500 instance pairs per CoInCo lemma, when available.   
We balance the two classes (T and F) and 
merge the obtained instances with \newcite{gari-soler-etal-2019-word}'s dataset (5,023 pairs) removing the duplicates. In total, we have 22,226 CoInCo instance pairs  for fine-tuning. We use these instances in combination with the Usim data.

\paragraph{Word-in-Context (WiC)} 
The WiC dataset contains 
pairs of  word instances in context  with the same or a different meaning   \cite{PilehvarandCamachoCollados}. 
Sentences come from WordNet \cite{Fellbaum1998}, VerbNet \cite{KipperSchuler2006} and Wiktionary examples, and were automatically annotated based on information provided in these resources.
We use the training set (5,428 sentence pairs) with its labels (T or F) as data for fine-tuning.

\begin{table*}[t!] 
\centering
\scalebox{0.9}{
\begin{tabular}{m{0.8cm}| m{7.8cm} m{7.8cm}}
\hline
   Class & Sentence 1 & Sentence 2 \\
   \hline
   
   
   
   \multicolumn{3}{c}{Usim} \\
  \hline
  
  \parbox{0.7cm}{\begin{center} T (4.3) \end{center}} &  We recommend that you \textbf{check} with us beforehand. & I have \textbf{checked} multiple times with my order and that is not the case.\\ 


\parbox{0.7cm}{\begin{center} F (1.3) \end{center}} & The romance is uninspiring... and \textbf{dry}. &  If the mixture is too \textbf{dry}, add some water; if it is too soft, add some flour. \\ 


  \hline
  \multicolumn{3}{c}{CoInCo} \\
  \hline

\parbox{0.7cm}{\begin{center} T \end{center}} & A mission to end a \textbf{war} \{\textit{\underline{fight}, \underline{battle}, \underline{conflict}, \underline{combat}, struggle, crusade}\} & He knew the \textbf{war} would soon be over and he would be heading home. \{\textit{\underline{fight}, \underline{battle}, \underline{conflict}, \underline{combat}, bloodshed, fighting, hostility}\} \\ 

\parbox{0.7cm}{\begin{center} F \end{center}} & You're all \textbf{right}? \{\textit{ok, okay, well, safe, good}\} &  He's sitting \textbf{right} there at the bar! \{\textit{over, straight, exactly, direcly, just, precisely, currently}\} \\  

  \hline
   \multicolumn{3}{c}{WiC} \\
  \hline
   \parbox{0.7cm}{\begin{center} T \end{center}} & Laws limit the \textbf{sale} of handguns . & They tried to boost \textbf{sales}. \\
   \parbox{0.7cm}{\begin{center}F \end{center}} & She didn't want to \textbf{answer}. & 	This may \textbf{answer} her needs. \\
  \hline
  \multicolumn{3}{c}{ukWaC-subs} \\
  \hline

   \parbox{0.85cm}{\begin{center} a (T)  \end{center}}  & For neuroscientists, the message was \textbf{clear}. & For neuroscientists, the message was \textbf{unambiguous}. \\
    \parbox{0.85cm}{\begin{center} b (F) \end{center}}  &  Need a \textbf{present} for someone with a unique name? &  Need a \textbf{moment} for someone with a unique name? \\
    \parbox{0.85cm}{\begin{center}c (F') \end{center}}  & Overdue tasks display on the due \textbf{date}. & Overdue tasks display on the due \textbf{heritage}. \\
\hline
  \multicolumn{3}{c}{Opusparcus} \\
  \hline
   \parbox{0.7cm}{\begin{center}T \end{center}}  & I \textbf{love} you so much & I \textbf{love} you to the moon and back. \\
  \parbox{0.7cm}{\begin{center}F \end{center}}  & yes, Mary, I would \textbf{love} to dance. & Why do I \textbf{love} him? \\

\hline

\end{tabular}}
\caption{Examples 
of instance pairs from each dataset used for fine-tuning. 
} \label{tab:dataset_examples} 
\end{table*}

\paragraph{ukWaC-subs}
The GWSC task 
addresses pairs of different words that can have similar meanings in some contexts and not in others (e.g., {\it room} and {\it cell}). 
Given that no training data is available, we automatically create one more dataset for fine-tuning called ukWaC-subs, which approximates this task.

ukWaC-subs contains pairs of sentences $(p_1, p_2)$ that differ in one word only. We create data 
by substituting a word $w$ in $p_1$ by  either 
(a) a correct substitute; 
(b) a word that is a good synonym of $w$ and could have been a correct substitute in another context but not in this one; 
or (c) a random word of the same part of speech as $w$. 
This is illustrated by the three ukWaC-subs sentences in Table  \ref{tab:dataset_examples}. With (a), we expect BERT to learn that \textit{clear} is being used in its \textit{unambiguous} sense in this context. In (b), we tell BERT that despite the (out-of-context) similarity between \textit{present} and \textit{moment}, the latter is not adequate in this context. With (c),  
we 
want BERT to learn to distinguish \textit{date} from a completely unrelated word  (\textit{heritage}). We use this data for a 3-way classification task.

We create this dataset 
by gathering sentences from the ukWaC corpus \cite{baroni2009wacky} and automatically annotating them with lexical substitutes. We identify the content words in 
a sentence and 
use as their candidate substitutes their paraphrases in 
the Paraphrase Database (PPDB) lexical XXL package \cite{ganitkevitch2013ppdb,pavlick2015ppdb}.\footnote{\url{http://paraphrase.org/}} The PPDB resource 
was automatically constructed 
by a  bilingual pivoting method. Every paraphrase pair 
has a PPDB 2.0 score indicating its quality. We only consider as candidates for substitution  
pairs with a score above 2. 
We then use the context2vec lexical substitution model  \cite{melamud2016context2vec} to rank the candidates 
according to how well they fit in a context. 
context2vec is a biLSTM model that jointly learns static representations of words and dynamic context representations. We rank candidate substitutes using the following formula:
\begin{equation} 
c2v\_score = \frac{cos(s,t) + 1}{2} \times \frac{cos(s,C) + 1}{2}
\label{c2v_formula}
\end{equation}

\noindent where $s$ is the static representation of the candidate substitute, $C$ is the context embedding of the sentence and $t$ is the static embedding of a word instance $i$ 
we want to replace. 
Using this formula, we obtain an ordered ranking $R$ of substitutes for an  instance $i$ in context $C$. The highest-ranked substitute is viewed as correct 
and 
serves to create instances of type (a). A random word of the same part of speech found in the corpus makes an instance of class (c). To obtain instances of class (b) we could in principle take the last substitute in the ranking. However, due to the noise that exists in PPDB, 
these often are not correct paraphrases of the target word. 
We therefore apply a filtering strategy proposed by \newcite{gari-soler-etal-2019-word} which checks 
whether substitutes in adjacent positions $(s_i, s_{i+1})$ in the ranking $R$ form a paraphrase pair in PPDB. If this is not the case for a specific pair, we stop checking and 
retain $s_{i+1}$ as a substitute that represents a different meaning of the target word. 

Once the substitutes have been collected, 
40\% of the instances are assigned to class (a), 30\% to class (b) and 30\% to (c). One sentence may contain more than one training instance if a substitute ranking is available for different words in it. 
A training instance is created by replacing the word with the 
substitute required by 
the class it has been assigned to. 
We 
create 100,000 
instances that we use to fine-tune BERT.

\paragraph{OpusParcus} 

 


\newcite{shi-etal-2019-retrofitting} show that retrofitting ELMo with paraphrases improves its performance on lexical semantic tasks. We follow a similar approach and use paraphrases 
to fine-tune BERT 
before applying it to GWSC. 
We 
use paraphrases from the Open Subtitles Paraphrase Corpus (Opusparcus) \cite{creutz-2018-open}. 
We use this corpus instead of the Microsoft Research Paraphrase Corpus \cite{dolan-etal-2004-unsupervised} used by  \newcite{shi-etal-2019-retrofitting}, because it  
contains paraphrase pairs for six European languages, including English and Finnish which are addressed in GWSC.  

Paraphrase pairs in Opusparcus 
were extracted from movies and TV  shows subtitles, and are ranked by quality. We use paraphrases 
from the Opusparcus training set 
with a quality score higher than 15,\footnote{Scores range from 
$\sim$77 (best quality) to 
$\sim$2 (worst quality).} and create our own training instances following the procedure of \newcite{shi-etal-2019-retrofitting}. 
Every pair of paraphrases that share a content word 
constitutes a positive example (T).  
For every T, we create a negative example (F) by selecting a pair of sentences from the resource 
that share the same word but are not paraphrases of each other. 
To avoid 
creating examples for target words that are highly frequent and have fuzzy semantics, we omit instances of the 200 most frequent 
words 
in the Google Books NGram corpus \cite{michel2011quantitative} 
(e.g., \textit{make}, \textit{get}, \textit{good}). 
In total, we use 100,000 sentence pairs for fine-tuning the English model and 60,520 for Finnish.

\subsection{Models}

We use these five datasets to fine-tune pre-trained BERT models for English and Finnish. 
All tasks require comparing the meaning of word instances in two different sentences.
We form an input sequence (sentence pair) for BERT by joining the two sentences together with the 
separator token ({\tt [SEP]}) in between.
Since the task is at the word level, we do not build our classifier on top of the {\tt [CLS]} token which is an aggregation of the whole input sequence. Instead, our classifier receives as input the BERT representations of the target word instances at the last layer. 
BERT uses wordpiece tokenization \cite{wu2016googles} which means that a target word may be split into several tokens. 
For words that have been split, we average the representations of each wordpiece. We use two 
kinds of heads for fine-tuning.

\begin{itemize}
\item {\bf Classification head}: The representations of the two target tokens are concatenated and fed to a linear classifier which outputs probabilities for each class. We use a cross entropy loss for training. We call this head {\sc classif}.

\item {\bf Cosine Distance head}: We apply the Cosine Embedding Loss (PyTorch \cite{pytorch2019}) to the representations of the two target tokens at the last layer. This loss increases the cosine distance of two tokens if they do not have the same meaning, and decreases it otherwise. We refer to this head as {\sc cosdist}.
\end{itemize}

\vspace{2mm}

\noindent Note that the ukWaC-subs dataset is compatible with the {\sc classif} head only because it has three classes.
 To predict the similarity of two target tokens in the GWSC task, we extract their representations 
from the different layers of a fine-tuned model. We use cosine similarity ($cossim$) as our similarity metric. 
In Subtask 2, which consists in predicting the similarity scores for a pair of words ($w_{a}$, $w_{b}$) in the same context, we simply calculate the cosine similarity of their representations in a specific layer.  
In Subtask 1, we need to predict a change in similarity between two words $w_{a}$ and $w_{b}$ 
in two different contexts ($c_{1}$, $c_{2}$). We estimate the change in similarity ($\Delta Sim$) with a simple subtraction:

\begin{equation}
\Delta Sim = cossim(w_{a_{c2}}, w_{b_{c2}}) - cossim(w_{a_{c1}}, w_{b_{c1}})
\end{equation}

\noindent where $w_{a_{c2}}$ is the representation of word $w_{a}$ in context $c_{2}$.

\subsection{Experimental Setup}

We participate in Subtasks 1 and 2 for  
English and Finnish.\footnote{We did not address Croatian and Slovenian due to the lack of datasets that could be used for fine-tuning.} For English, we fine-tune the {\tt bert-base-uncased} model. For Finnish, we use the uncased Finnish model ({\tt finnish}) \cite{virtanen2019multilingual}\footnote{\url{https://github.com/TurkuNLP/FinBERT}} and the uncased Multilingual BERT-base model ({\tt multilingual}).\footnote{\url{https://github.com/google-research/bert/blob/master/multilingual.md}} For faster fine-tuning, we set the maximum length to 128 wordpiece and omit 
examples where a target word occurs after this position. 

As a development set for English, 
we use the officially released GWSC trial data (10 sentence pairs) 
and an earlier release of trial data (8 sentence pairs), both distinct from  the test set. 
We use these data to select  
the best models and hyperparameters for our official submissions to GWSC. 
The English test set consists of 
340 context pairs for Subtask 1 
and 680 unique contexts for Subtask 2. 
We fine-tune {\tt bert-base-uncased} separately on each of our English datasets experimenting with the two classification heads \{{\sc classif}, {\sc cosdist}\} and with different learning rates \{5e-5, 1e-6, 1e-7\} for up to 15 epochs. 
These hyperparameters, along with the layer the word representations are extracted from, are set on the GWSC trial data. 
Our submitted models were fine-tuned on WiC, Opusparcus and CoInCo-Usim with a learning rate of 5e-5 and 0.1 dropout for 4, 3 and 2 epochs, respectively.
The ukWaC-subs model was fine-tuned for 11 epochs with a learning rate of 1e-6 and 0.2 dropout. Dropout was determined based on results on 2,000 held-out ukWaC-subs instances.
Since no trial dataset was released for Finnish, we fixed the hyperparameters for our models to those that worked best for the English Opusparcus data. Our submitted predictions are from the higher layers of the models fine-tuned with the {\sc classif} head.
The test set for Finnish consists of 24 context pairs for Subtask 1 (48 unique contexts for Subtask 2).\footnote{We use the HuggingFace \texttt{transformers} library \cite{Wolf2019HuggingFacesTS} to implement our experiments.}


The metrics used to evaluate model predictions are the uncentered Pearson correlation ($\rho$) in Subtask 1,  
and the harmonic mean of Pearson and Spearman correlations ($\bar\rho$) in Subtask 2.

\begin{table}[t]

    \centering
    \begin{tabular}{lcc}
    
         Model & Subtask 1 & Subtask 2\\
         \toprule 
         \multicolumn{3}{c}{English} \\
         \toprule 
          WiC {\sc cosdist} & $\dagger$ 0.760$_{11}$ & 0.689$_{11}$ \\
         ukWaC-subs & 0.751$_{10}$ & $\dagger$ \textbf{0.718}$_{10}$ \\ 
         Opusparcus {\sc classif} & $\dagger$ 0.751$_{11}$  & 0.669$_{6}$ \\ 
         CoInCo + Usim {\sc cosdist} & \textbf{0.765}$_{10}$ & $\dagger$ 0.686$_{6}$ \\ \hline 
         {\tt bert-base-uncased} & 0.715$_{11}$ & 0.661$_{11}$ \\
          \toprule
         \multicolumn{3}{c}{Finnish} \\
         \toprule
         {\tt multilingual} Opusparcus {\sc classif}  & $\dagger$ 0.593$_{9}$ & $\dagger$ 0.192$_{11}$ \\ 
         {\tt multilingual} Opusparcus {\sc classif}  & \textbf{0.718}$_{6}$ &  0.286$_{5}$  \\ 
         {\tt finnish} Opusparcus {\sc classif}  & $\dagger$ 0.500$_{12}$ & $\dagger$ 0.491$_{9}$  \\
         {\tt finnish} Opusparcus {\sc classif}  & 0.550$_{1}$ & 0.568$_{3}$  \\ \hline
         
         {\tt multilingual} & 0.677$_{11}$ &  0.388$_{9}$ \\
         {\tt finnish} & 0.577$_{12}$ & \textbf{0.671}$_{12}$  \\
    \end{tabular}
    \caption{Results of our English and Finnish models in GWSC Subtasks 1 and 2. 
    The models are compared to three BERT-based baselines without fine-tuning. 
    The evaluation metric in Subtask 1 is Pearson's correlation coefficient. In Subtask 2, it is the harmonic mean of Pearson and Spearman's correlation coefficients. Our  official submissions to the GWSC task for each language are marked with $\dagger$. Subscripts indicate the BERT model layer used.}
    \label{tab:results}
\end{table}{}

\section{Results}

Results for the two English and Finnish subtasks 
are presented in Table \ref{tab:results}. We report  
results of the two best systems submitted to each subtask (marked with $\dagger$) along with results calculated during the post-evaluation phase for comparison. These include 
baseline predictions  made by BERT models without fine-tuning. 

Although the two subtasks are highly related, different models perform best  
in each one.
For English, the best result 
in Subtask 1 (among our official submissions) 
is obtained by the model fine-tuned on WiC data with the {\sc cosdist} head ($\rho=0.760$) which occupies the third position in the final ranking. It is closely followed by the model fine-tuned on paraphrase data with the {\sc classif} head.
The best performing model in Subtask 2  is the one 
fine-tuned on the ukWaC-subs data ($\bar\rho=0.718$) which ranked fourth. The second best model 
uses the {\sc cosdist} head and 
is trained on the CoInCo and Usim data together. 
All English models outperform the  BERT-based baseline without fine-tuning ($\rho=0.715$ and $\bar\rho=0.661$).  
This demonstrates 
the higher quality of lexical semantic knowledge in our fine-tuned models.

Best results for the Finnish  Subtasks 1 and 2 
are also 
produced by different models. 
The {\tt multilingual} model performs better on Subtask 1 and the {\tt finnish} model 
on Subtask 2. 
We observe that the {\tt multilingual} model tends to assign very high similarities to all word instance pairs, which explains its low performance in Subtask 2. At the same time, however, it does well on Subtask 1 because it captures the magnitude of the difference in similarity between two pairs.
Given that no trial data (development set) are available for Finnish and that the maximum number of submissions was nine, we could only try at most five layers per model 
at submission time. 
The models were ranked sixth and fourth in Subtasks 1 and 2.

During the post-evaluation phase, we had the possibility to test all layers of the models. The sixth layer of the {\tt multilingual} model fine-tuned on Finnish Opusparcus data outperforms the {\tt multilingual} baseline on Subtask 1 ($\rho=0.718$ vs $\rho=0.677$), but the other fine-tuned models do not improve over their respective baselines. 
Surprisingly, the {\tt finnish} baseline model 
in Subtask 2 ($\bar\rho=0.671$) 
outperforms the top-ranked model for Finnish among all teams that participated in the task ($\bar\rho=0.645$). 


\section{Conclusion}

We have participated in the SemEval task Graded Word Similarity 
in Context for English and Finnish, with  models integrating different notions of word similarity. We have specifically investigated the effect of fine-tuning pre-trained BERT models on existing datasets that address word meaning similarity in context. Furthermore, we have proposed a new fine-tuning task where in-context lexical similarity is approximated through automatic substitute annotations.

Our English models are ranked at the third and fourth position in the two GWSC subtasks, and outperform a BERT-based baseline without fine-tuning. 
This demonstrates the benefit of fine-tuning BERT on a task that is closely  
related to the end task, even when the data used for fine-tuning are automatically obtained. Due to the scarcity of resources for Finnish, we could only fine-tune models with paraphrases. The Finnish models are mid-ranked among all participating systems.

\newpage

\section*{Acknowledgements}


\setlength\intextsep{0mm}
\begin{wrapfigure}[4]{l}{0pt}
\includegraphics[scale=0.3]{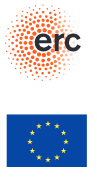}
\end{wrapfigure}

This  work has been supported by the French National Research Agency under project ANR-16-CE33-0013. The work is also part of the FoTran project, funded by the European Research Council (ERC) under the European Union’s Horizon 2020 research and innovation programme (grant agreement \textnumero ~771113).

\bibliographystyle{coling}
\bibliography{semeval2020}

 



\end{document}